\documentclass[a4paper,fleqn]{cas-sc}

\usepackage[authoryear]{natbib}
\usepackage{amsmath,amssymb,booktabs,placeins}
\usepackage{float}
\newcommand{\degree}{^{\circ}}
\newcommand{\figimage}[1]{\includegraphics[width=\linewidth,keepaspectratio]{figures/#1}}

\def\tsc#1{\csdef{#1}{\textsc{\lowercase{#1}}\xspace}}
\tsc{WGM}
\tsc{QE}

\begin{document}
\let\WriteBookmarks\relax
\def\floatpagepagefraction{1}
\def\textpagefraction{.001}

\shorttitle{Vision Foundation Model Adaptation for Fracture Characterization}    

\shortauthors{G. Wang et~al.}  

\title [mode = title]{Adapting a Vision Foundation Model for Fracture Characterization in Acoustic Image Logs}  



%

\author[1, 2]{Guangyu Wang}[orcid=0000-0002-7932-0407]
\ead{wangguangyu@mail.ustc.edu.cn}
\credit{Writing – original draft, Visualization, Methodology, Formal analysis, Data curation}

\author[1, 2]{Xiaodong Ma}[orcid=0000-0003-3373-808X]
\ead{maxiaodong@ustc.edu.cn}
\credit{Writing – review \& editing, Validation, Supervision, Resources, Methodology, Investigation, Funding acquisition, Formal analysis, Conceptualization}
\cormark[1]

\author[1, 2]{Xinming Wu}[orcid=0000-0002-4910-8253]
\ead{xinmwu@ustc.edu.cn}
\credit{Writing – review \& editing, Validation, Supervision, Methodology, Investigation, Formal analysis, Conceptualization}
\cormark[1]



\affiliation[1]{organization={State Key Laboratory of Precision Geodesy, School of Earth and Space Sciences, University of Science and Technology of China},
            city={Hefei},
            postcode={230026}, 
            country={China}}

%
%
%
%

\affiliation[2]{organization={Mengcheng National Geophysical Observatory, University of Science and Technology of China},
            city={Mengcheng},
            postcode={233500}, 
            country={China}}

\cortext[1]{Corresponding author}



\begin{abstract}
Natural fractures (NFs) and drilling-induced tensile fractures (DITFs) observed in acoustic image logs provide important information about rock mass structure and in situ stress conditions. However, characterizing their locations and orientations requires time-consuming expert interpretation. Deep learning offers a promising way to automate this process, but its application is constrained by the scarcity of expert-annotated fracture data. Pretrained vision foundation models (VFMs) may help address this limitation by providing transferable visual representations that can be adapted with relatively few task-specific annotations. In this study, we investigate whether adapting a pretrained VFM improves fracture characterization compared with training task-specific models from scratch on the same limited labeled dataset. Using only 80 annotated acoustic image samples, we adapted a DINOv3 model pretrained on satellite imagery for fracture characterization and evaluated it using data from four boreholes independent of the fine-tuning dataset. The fine-tuned DINOv3 model characterized NF and DITF parameters at a level comparable to human interpretation. Compared with U-Net, YOLOv8, and Swin Transformer models trained from scratch, the fine-tuned DINOv3 produced fewer false positives and false negatives and more accurate estimates of fracture counts and orientations. However, experiments with progressively fewer training samples indicate that DINOv3’s advantage over the task-specific models generally diminishes as the number of training/fine-tuning samples decreases.  We also examined how decoder design, fine-tuning strategy, and image enhancement affect fracture characterization performance. Among the configurations examined, a progressive upsampling decoder combined with full fine-tuning yielded the best DINOv3 performance, whereas image enhancement intended to emphasize fracture appearance did not necessarily improve fracture characterization accuracy. These findings support VFM adaptation as an effective strategy for characterizing fractures in acoustic image logs when labeled data are limited. The proposed workflow may also provide a useful reference for applying VFMs to other geoscientific image analysis tasks with limited annotations.
\end{abstract}




\maketitle

\section{Introduction}
Understanding subsurface geological features is a fundamental objective in georesource development and geoscientific investigation. Among the features accessible at the borehole scale, fractures are of particular importance. Natural fractures (NFs) record the deformation history of the rock mass \citep{wenning2017image,massiot2017evidence} and impose a major influence on its mechanical and hydraulic properties \citep{massiot2017statistical,moein2018fracture}. Drilling-induced tensile fractures (DITFs) develop in response to stress concentration around the borehole wall and provide constraints on stress orientation \citep{brudy1999drilling,zoback2003determination}. Accurate characterization of fracture locations and orientations is therefore essential for hydrocarbon reservoir evaluation \citep{milad2018upscaled,lai2021fracture}, enhanced geothermal system development \citep{valley2009stress,chen2026fracture}, and in situ stress assessment \citep{nian2016stress,shahri2016stress}.

\begin{figure}
  \centering
  \centering\figimage{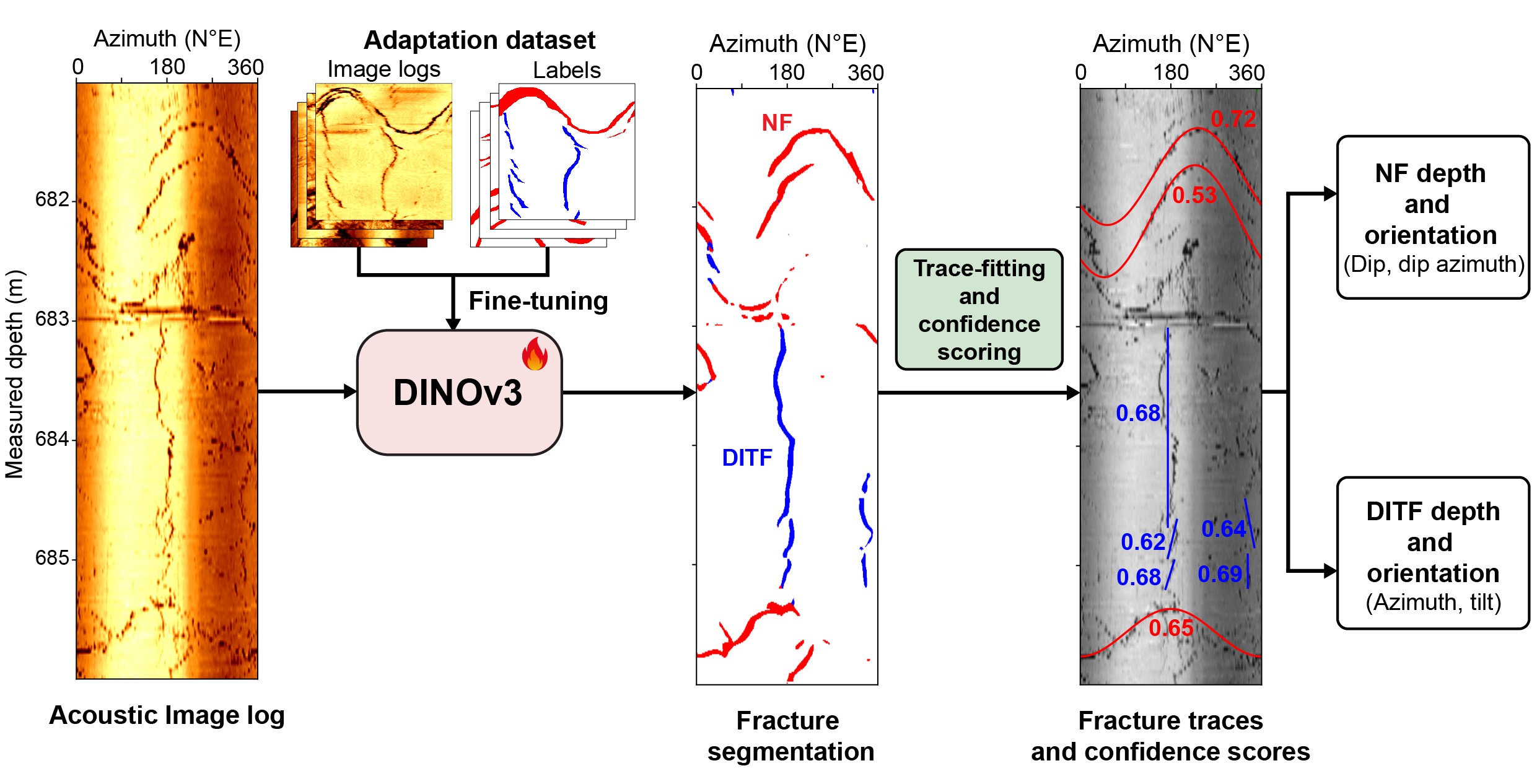}
  \caption{Workflow for fracture characterization through DINOv3 adaptation.NF denotes natural fracture and DITF represent drilling-induced tensile fracture. The number beside each fracture trace represent the confidence score, ranging from 0 to 1. Fracture traces with confidence score less than 0.5 are not visualized.}
\label{fig:workflow}
\end{figure}

Acoustic image logs provide direct visualization of fractures on borehole walls. An acoustic image log is acquired through emitting focused ultrasonic pulses toward the borehole wall and recording the amplitude and travel time of the reflected signal. In this study, acoustic image log only refers to its amplitude component. As shown in Figure \ref{fig:workflow}, NFs typically appear as sinusoidal traces whose geometry encodes the fracture orientation.

DITFs commonly appear as axial or en echelon traces, often occurring in pairs approximately 180° apart in azimuth. The azimuth of DITFs reflects the maximum horizontal stress ($S_{\mathrm{Hmax}}$) direction. Conventional fracture characterization involves identifying these fracture traces and fitting their geometry manually. The repetitive nature of this work and the considerable time it requires have driven the development of automated methods.

Early attempts at automated fracture characterization from borehole images employs classical image processing techniques, including edge detection operators \citep[e.g.,][]{hall1996automatic, wang2005edge}, Hough and Radon transforms for sinusoidal curve detection \citep[e.g.,][]{glossop1999implementation, kherroubi2008automatic}, mathematical morphology \citep{xavier2015fracture}, and phase congruency methods \citep{assous2014automated}. While these approaches offered improved efficiency over manual picking, they generally struggled with the highly variable image quality, complex geological backgrounds, and diverse morphological expressions of fractures encountered in real-world logging data \citep{han2023automatic}.

The development of deep learning has brought significant advances in automated fracture characterization. Convolutional neural networks (CNNs), particularly the U-Net architecture and its variants, have been widely adopted for pixel-level fracture segmentation from borehole image logs \citep[e.g.,][]{cruz2017improving,  li2025EILNet}. Object detection frameworks such as Faster R-CNN, Mask R-CNN, and YOLOv8 have also been applied to fracture localization and instance segmentation \citep[e.g.,][]{dias2020automatic, liu2022automatic, han2025intelligent}. More recently, transformer-based architectures have been explored for fracture segmentation \citep{wang2023fracture} and direct estimation of fracture locations and orientations \citep{nasim2025automated}, leveraging their capacity to model long-range dependencies in image features. 

Despite these advances, training deep learning models for image log interpretation remains challenging. Annotating fractures takes substantial time, so relatively few labeled acoustic images are available for training. Models trained from scratch on such small datasets may struggle to recognize fractures in new boreholes. Previous studies have addressed this problem through unsupervised domain adaptation, transferring information from labeled road crack images to unlabeled borehole images \citep{zhang2021fracture,zhang2022fracture}. These studies demonstrate the potential value of knowledge learned from other image types. We therefore investigate whether a pretrained vision foundation model can be adapted with a small number of labeled acoustic images to improve fracture characterization.

Vision foundation models (VFMs) offer a fundamentally different paradigm for addressing the label scarcity challenge. These large-scale neural networks, pretrained on massive image collections through self-supervised learning, develop rich and generalizable visual feature representations that can be effectively transferred to diverse downstream tasks with minimal task-specific adaptation \citep{bommasani2021opportunities}. The DINO family of models \citep{caron2021dino, oquab2024dinov2, simeoni2026dinov3} exemplifies this approach. By learning visual representations through discriminative self-supervised objectives on datasets comprising hundreds of millions to billions of natural images, these models acquire hierarchical feature representations that capture both global semantic context and local details without any human annotation. These representations have proven transferable to downstream tasks across diverse specialized domains. In medical image segmentation, fine-tuned DINOv2 consistently outperforms fully supervised U-Net baselines under limited annotation conditions, with the performance gap widening as training data decrease \citep{kundu2025assessing}. In industrial anomaly detection, DINOv2 achieves state-of-the-art one-shot defect localization on standard benchmarks without domain-specific fine-tuning \citep{damm2025anomalydino}. In geophysical image analysis, DINOv2 fine-tuned with parameter-efficient methods outperforms task-specific U-Net models on seismic facies classification, fault detection, and crater identification tasks, even with as few as 50 training samples \citep{guo2025cross}. Together, these results indicate that pretrained visual representations generalize effectively across imaging domains far removed from the natural image distribution on which they were trained.

In this study, we adapt a DINOv3 encoder pretrained on satellite imagery using a small set of labeled acoustic images (Figure \ref{fig:workflow}). The adapted model segments the images into background, NF, and DITF classes. Geometric processing then converts the predicted masks into fracture depths, orientations, and morphology-based confidence scores. We compare this approach with U-Net, YOLOv8, and Swin Transformer models trained from scratch on the same labeled samples, evaluating both image segmentation and fracture parameters against manual interpretations. We also examine how the number of training samples, decoder design, fine-tuning strategy, and image enhancement affect performance. By comparing the approaches quantitatively with limited labeled data, this study assesses whether fine-tuning a VFM yields better fracture characterization than training task-specific models from scratch. The findings also inform the choice of decoder and fine-tuning strategy and may guide similar studies beyond the task of image log fracture characterization.

\section{Method}
\subsection{Network architecture and fine-tuning strategy}
\begin{figure}
\centering\figimage{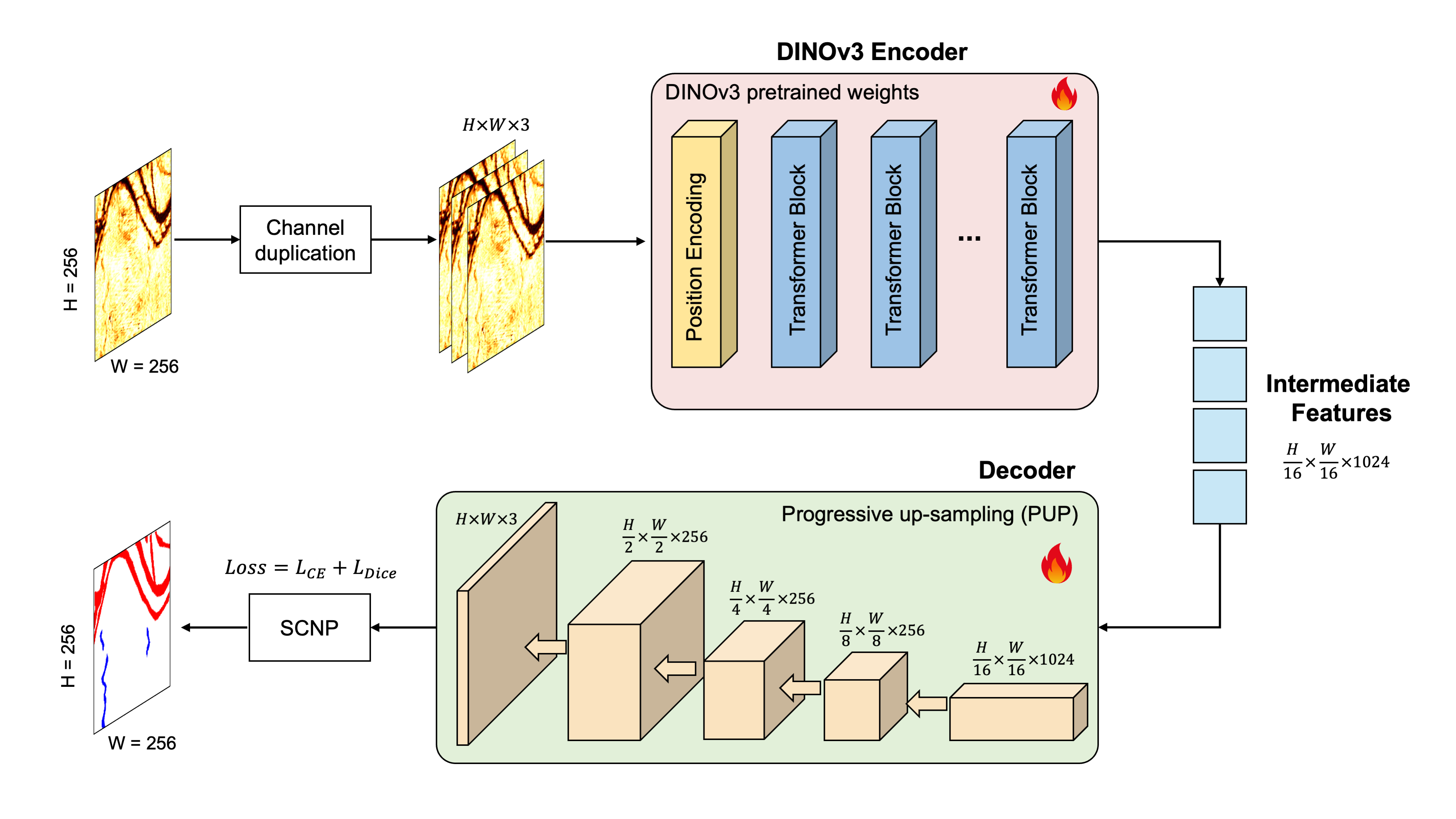}
\caption{Segmentation architecture and adaptation strategy. The pretrained DINOv3 encoder supplies spatial features to a progressive upsampling (PUP) decoder, which produces pixel-wise background, NF, and DITF predictions. Pretrained weights of DINOv3 are used to initialize the encoder, and both encoder and decoder are updated in during fine-tuning. The acoustic image log is replicated across the three input channels.}
\label{fig:architecture}
\end{figure}

Our workflow first identifies fracture pixels in an acoustic image and then calculates fracture parameters represented by those pixels. For the first step, we combine a pretrained DINOv3 image encoder, which extracts visual patterns, with a decoder that produces a pixel-level map (Figure \ref{fig:architecture}). The encoder is a vision transformer initialized from SAT-493M satellite imagery \citep{simeoni2026dinov3}. The decoder assigns each pixel in a $256\times256$ image to background, NF, or DITF class.

Our main configuration uses a progressive upsampling (PUP) decoder, which gradually expands the encoder's image features into a full-resolution fracture map \citep{zheng2021rethinking}. We update both the pretrained encoder and the decoder using labeled acoustic images. This fine-tuning strategy is referred to as full fine-tuning or \emph{Unfrozen} in the comparisons below. Additional comparisons of decoder designs and fine-tuning strategies are presented in Section~\ref{sec:adaptation}.

The acoustic amplitude values are standardized, and the single image channel is copied three times to match the encoder's input format. During training, we apply augmentation techniques to the input image logs, including to reverse images along depth, reflect them along azimuth, and shift them around the azimuthal image boundary. Each transformation is also applied to its label. These transformations create varied training views but do not add new, independently labeled samples. The training combines cross-entropy and Dice losses to compare predicted fracture regions with the labels. It also uses Same Class Neighbor Penalization (SCNP) to discourage breaks in narrow fracture masks \citep{valverde2026scnp}.The network is trained for 500 epochs using AdamW optimization method. When processing a continuous log, we divide it into overlapping images along depth and combine predictions in the shared regions. A 20\% overlap is applied to help avoid artificial breaks at image boundaries. 

We compare DINOv3 with three models trained from scratch. These are U-Net \citep{ronneberger2015unet}, a YOLOv8-based segmentation model \citep{jocher2023yolov8}, and a Swin Transformer segmentation model \citep{liu2021swin}. The YOLOv8 model predicts fracture pixels rather than bounding boxes. All models use the same labeled samples, training duration, batch size, and SCNP setting.

\subsection{Extracting fracture depths and orientations}
The models described above identify NF and DITF pixels, but fracture characterization requires measurements of individual traces. We therefore process the two fracture types separately. In an image log, the vertical direction represents depth and the horizontal direction represents azimuth, or position around the borehole wall. We group connected fracture pixels into traces, extract a centerline for each trace, and convert its image positions to measured depths and azimuths.

On an image log, an NF typically appears as a sinusoidal trace. We fit a sinusoid to its centerline. The fitted curve’s vertical midpoint gives the fracture’s central depth. Its amplitude, together with the borehole diameter, determines the fracture dip. The azimuth of the sinusoid's deepest point indicates the dip azimuth of the fracture.We also assign each NF a confidence score, giving 70\% weight to how far the trace extends around the borehole and 30\% to how closely it follows the fitted curve. Thus, a trace that covers more of the borehole circumference and fits the curve more closely receives a higher score. The confidence score reflects how closely a detected trace matches the expected fracture geometry. A higher score indicates a more credible fracture pick, whereas a lower score suggests that the trace may be noise.

In contrast, DITFs often appear as nearly vertical or en echelon traces. We fit a straight line to the overall trend of each DITF centerline and use it to estimate the trace’s central depth, azimuth, length, and tilt relative to the borehole axis. We also assign each DITF a confidence score. Because DITFs often extend along depth and occur in pairs on approximately opposite sides of the borehole, the score gives 30\% weight to depth persistence and 70\% to evidence of a counterpart at a similar depth. The pairing assessment considers both azimuthal separation from $180\degree$ and depth overlap. Under this confidence scoring criterion, some short, unpaired false-positive noise features receive relatively low confidence scores. However, a real DITF may nevertheless receive a low score if its counterpart is faint or absent.

\subsection{Dataset and evaluation design}
\begin{figure}
\centering\figimage{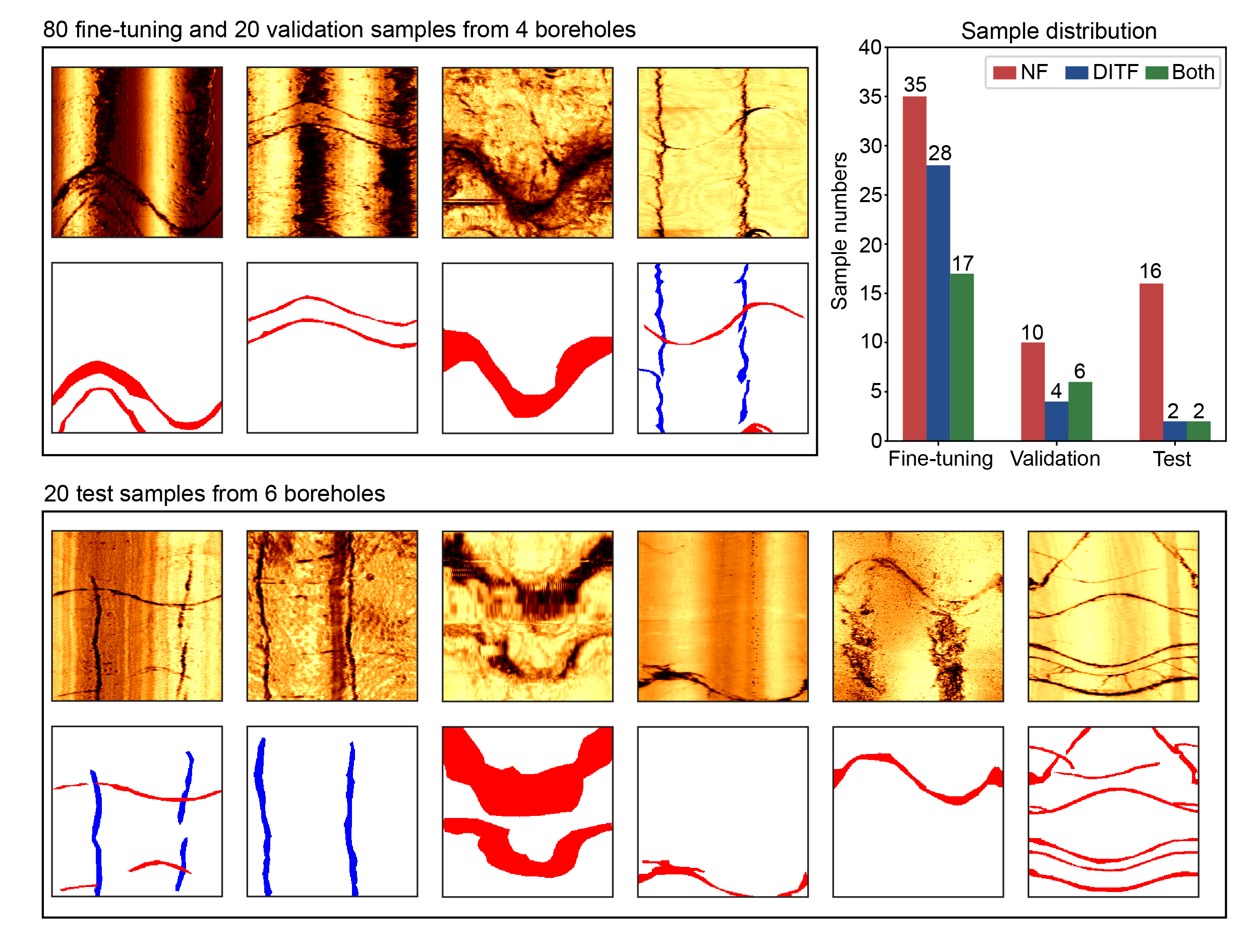}
\caption{Acoustic image log dataset used for DINOv3 fine-tuning and evaluation. Representative images show the range of fracture expressions and backgrounds, with one example per borehole. Fine-tuning and validation images originate from four boreholes, and test images from six other boreholes. Histograms summarize the numbers of NF-only, DITF-only, and mixed images.}
\label{fig:dataset}
\end{figure}

The acoustic image logs used in this study were obtained from boreholes associated with geothermal projects, scientific drilling, and other subsurface investigations. They cover different rock types and imaging conditions and show a range of NF and DITF appearances. Each labeled sample is processed as a $256\times256$-pixel acoustic-amplitude image. NF and DITF regions are marked at the pixel level, with the remaining pixels assigned to background. Figure \ref{fig:dataset} shows representative images and their labels.

The dataset contains 120 labeled images. 80 images were used for training and 20 for validation, all from four boreholes. The remaining 20 images were reserved for testing and came from six other boreholes that were not used for training or validation. This separation allows us to assess segmentation performance on images from previously unseen boreholes. Figure \ref{fig:dataset} also shows how many images contain only NFs, only DITFs, or both. The validation and test sets include images containing only NFs, only DITFs, and combinations of both, providing a comprehensive evaluation of the model’s ability to segment different types of fractures.

We evaluate segmentation with intersection over union (IoU). For each image, we divide the number of pixels marked as fracture by both the model and the expert by the number marked as fracture by either one. We then report the mean across images. IoU measures how well the predicted fracture regions agree with the manual labels. However, good agreement at the pixel level does not necessarily mean that the resulting fracture depths and orientations are accurate. We therefore also evaluate the picks and extracted parameters for NFs and DITFs separately.

To assess whether the predicted masks lead to useful fracture measurements, we compare automated picks from continuous logs with manual interpretations. The following matching tolerances allow for small differences in position or orientation while still requiring the picks to describe approximately the same fracture. For NFs, central depths must differ by no more than 0.1 m and orientations by no more than $15\degree$. For DITFs, azimuths must differ by less than $30\degree$, and the depth intervals must overlap by more than 80\% of at least one trace's length. 

We use the false-negative rate (FNR) to measure the fraction of manual picks missed by the model, so a lower value indicates more complete detection. The false-positive rate (FPR) is used to reflect the fraction of automated picks with no manual match, so a lower value indicates fewer unmatched candidates. FNR follows its usual definition, but the reported FPR differs from the conventional false-positive rate because its denominator is the number of automated picks, not the number of negative cases. Manual picks serve as the reference for these comparisons, but it does not represent an error-free record.

We also compare summary properties in consecutive 5 m depth intervals. These metrics show whether the model reproduces the amount and prevailing orientation of fractures along a log, even when individual picks are imperfect. For NFs, we calculate mean absolute differences in fracture count, mean dip azimuth, and mean dip. For DITFs, we compare total trace length, mean azimuth, and mean tilt in the same way. A trace is assigned to the interval containing its central depth. Azimuths near $0\degree$ and $360\degree$ are treated as close directions, while DITF directions $180\degree$ apart are treated as equivalent. Geometric errors are calculated only where the required values are available. They must therefore be read alongside the missed- and unmatched-pick fractions.

To make fracture-picking accuracy easier to compare across methods, we combine detection and geometric accuracy in one summary score. Detection is represented by the fractions of manual picks recovered and automated picks matched. Geometry is represented by the three errors described above. Detection and geometry each contribute half of the score. To combine errors measured in different units, we compare each one with a fixed reference value. For each evaluated log $b$ with complete measurements, the score is
\begin{equation}
S_b=25\bigl[(1-\mathrm{FNR}_b)+(1-\mathrm{FPR}_b)\bigr]
+\frac{50}{3}\sum_{j=1}^{3}\exp(-e_{bj}/m_{bj}).
\label{eq:composite}
\end{equation}
Here $e_{bj}$ is geometric error $j$. Its reference value $m_{bj}$ is the median for that error across all compared methods and remains fixed throughout the analyses. NF and DITF scores are averaged separately across the logs evaluated for each fracture type. Higher scores indicate better overall picking performance.

\FloatBarrier
\section{Results}
In this section, we first compare how accurately the fine-tuned DINOv3 model and the models trained from scratch segment fractures in individual images. We then assess whether the predicted traces produce useful NF and DITF picks and fracture parameters in continuous logs from four boreholes not used for segmentation training or validation. Together, the image-level and log-level results show where improved segmentation translates into better fracture characterization and where errors remain.

\subsection{Fracture segmentation}
\begin{figure}
\centering\figimage{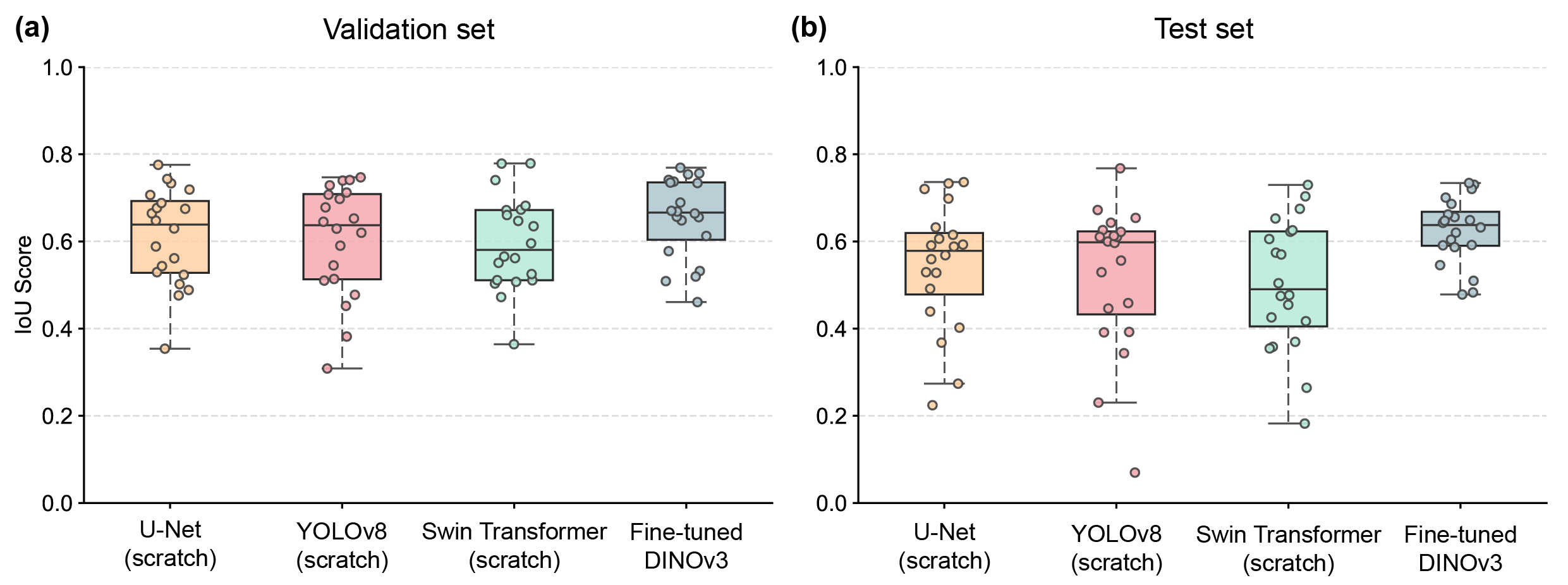}
\caption{IoU distributions of NFs and DITFs for (a) validation and (b) test images. U-Net, YOLOv8, and Swin Transformer are trained from scratch, whereas DINOv3 is fully fine-tuned. }
\label{fig:segmentation_metrics}
\end{figure}

For fracture segmentation, the fine-tuned DINOv3 achieves the highest mean IoU on both the validation and borehole-separated test sets (Figure \ref{fig:segmentation_metrics}a for validation, Figure \ref{fig:segmentation_metrics}b for testing, and Table~\ref{tab:iou}). On the test set, its mean IoU is 0.624, compared with 0.545 for U-Net, the strongest model trained from scratch. This is a relative improvement of 14.5\%. DINOv3 also leads on the validation set, with a mean IoU of 0.655 versus 0.611 for U-Net.

\begin{table}[tbp]
\caption{Mean IoU scores of NFs and DITFs for the validation and test dataset, respectively.}
\label{tab:iou}
\centering
\begin{tabular}{lcc}
\toprule
Model and learning strategy & Validation & Test\\
\midrule
U-Net, from scratch & 0.611 & 0.545\\
YOLOv8, from scratch & 0.604 & 0.522\\
Swin Transformer, from scratch & 0.597 & 0.502\\
DINOv3, full fine-tuning & \textbf{0.655} & \textbf{0.624}\\
\bottomrule
\end{tabular}
\end{table}

\begin{figure}
\centering\figimage{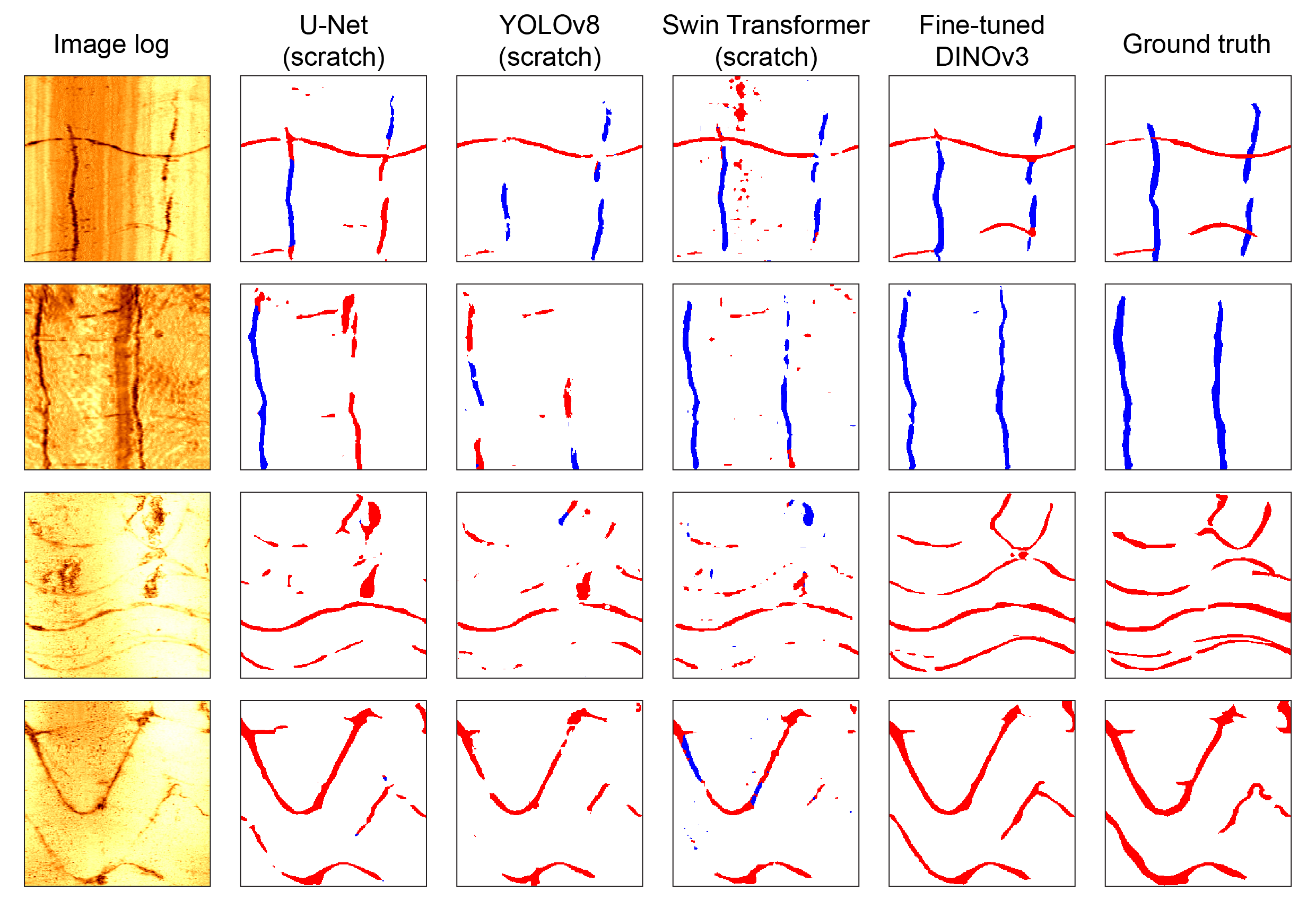}
\caption{Segmentation of four selected test images by the fine-tuned DINOv3 model and the three from-scratch baselines (U-Net, YOLOv8, and Swin Transformer). NF and DITF labels are shown in red and blue, respectively, with manual segmentation as the reference.}
\label{fig:segmentation_examples}
\end{figure}

All models perform less well on the test set than on validation, showing that segmentation remains sensitive to differences between boreholes. The mean IoU decreases for DINOv3 are less than the decreases for the models trained from scratch. This pattern is consistent with better transfer to the test boreholes in this comparison, although 20 test images cannot represent all logging conditions. Selected examples in Figure \ref{fig:segmentation_examples} show how these numerical differences appear in the predicted fracture traces.

In these examples, DINOv3 generally recovers more continuous annotated traces while producing fewer responses to background features than the comparison models. Errors remain around faint or intersecting traces and complex backgrounds. The examples illustrate the segmentation result but do not establish whether every predicted trace yields a correct fracture pick. We therefore examine the extracted NF and DITF parameters separately below.

\subsection{Natural fracture characterization}
\begin{figure}
\centering\figimage{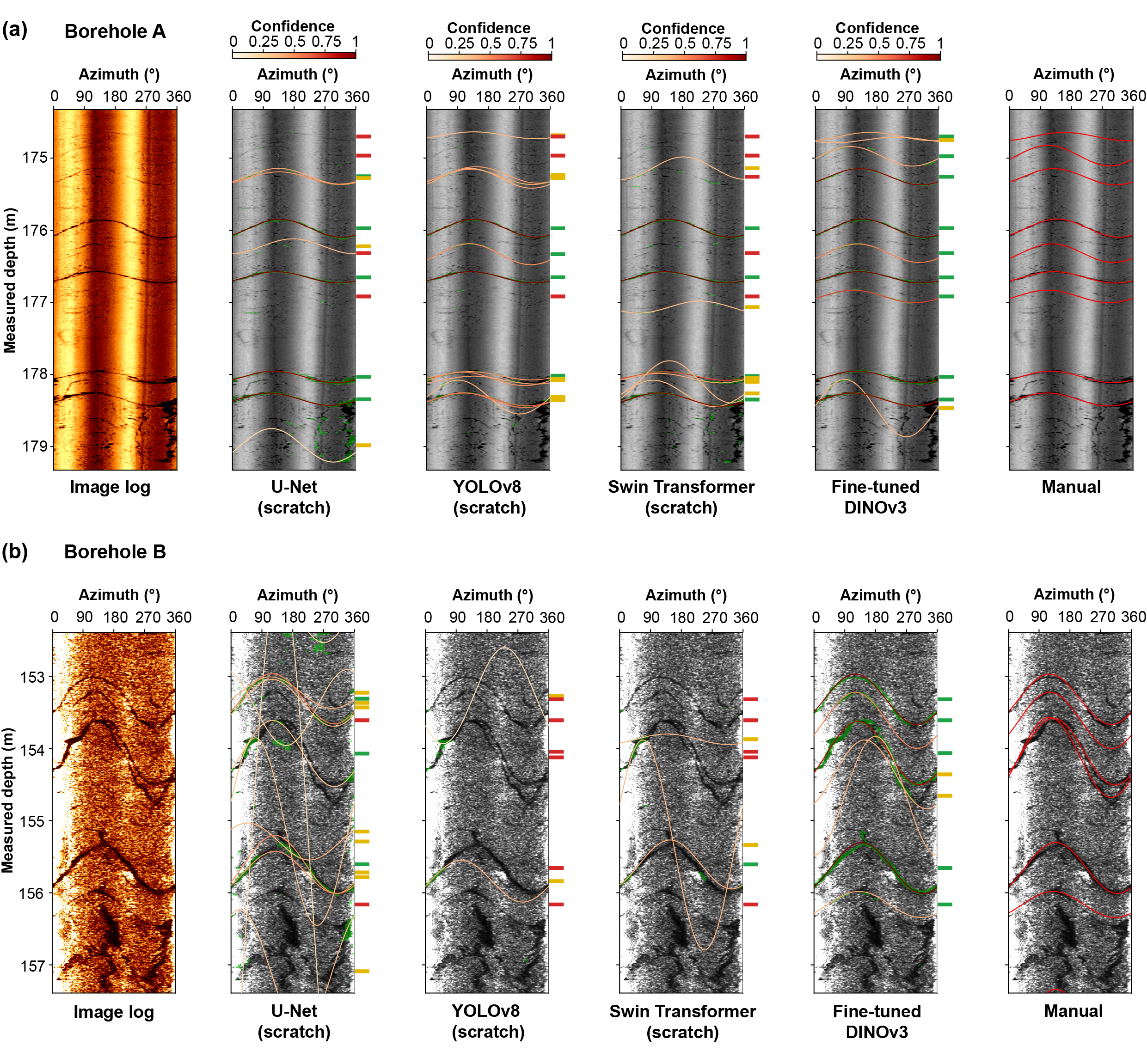}
\caption{NF picking in selected 5 m intervals of (a) Borehole A and (b) Borehole B. Green regions denote predicted NF masks, and red curves denote fitted traces, with intensity indicating confidence. Side markers identify missed manual picks (red), unmatched automated picks (yellow), and matched picks (green) under the depth and orientation tolerances.}
\label{fig:nf_examples}
\end{figure}

We evaluate NF picking along two continuous logs, designated Boreholes A and B. When reviewing a selected interval, confidence scores help interpreters identify fitted traces that more closely follow the expected NF geometry. In Figure \ref{fig:nf_examples}, darker curves represent higher confidence scores and paler curves represent lower scores, allowing interpreters to focus first on the more plausible candidates while checking lower-scoring picks against the image.

The selected intervals show how image quality affects the recovery of sinusoidal traces. In Borehole A (Figure \ref{fig:nf_examples}a), the relatively smooth borehole wall produces a less noisy image, although pronounced tool eccentering creates two dark vertical bands. DINOv3 follows several of the manually interpreted traces despite these bands, whereas the models trained from scratch miss more traces or add unmatched picks. In Borehole B (Figure \ref{fig:nf_examples}b), the rougher borehole wall produces a speckled image in which fracture traces are harder to distinguish from background texture. DINOv3 recovers more of the main interpreted traces in this example, although some of its picks remain unmatched. These contrasting examples motivate the full-log comparison below.

\begin{figure}
\centering\figimage{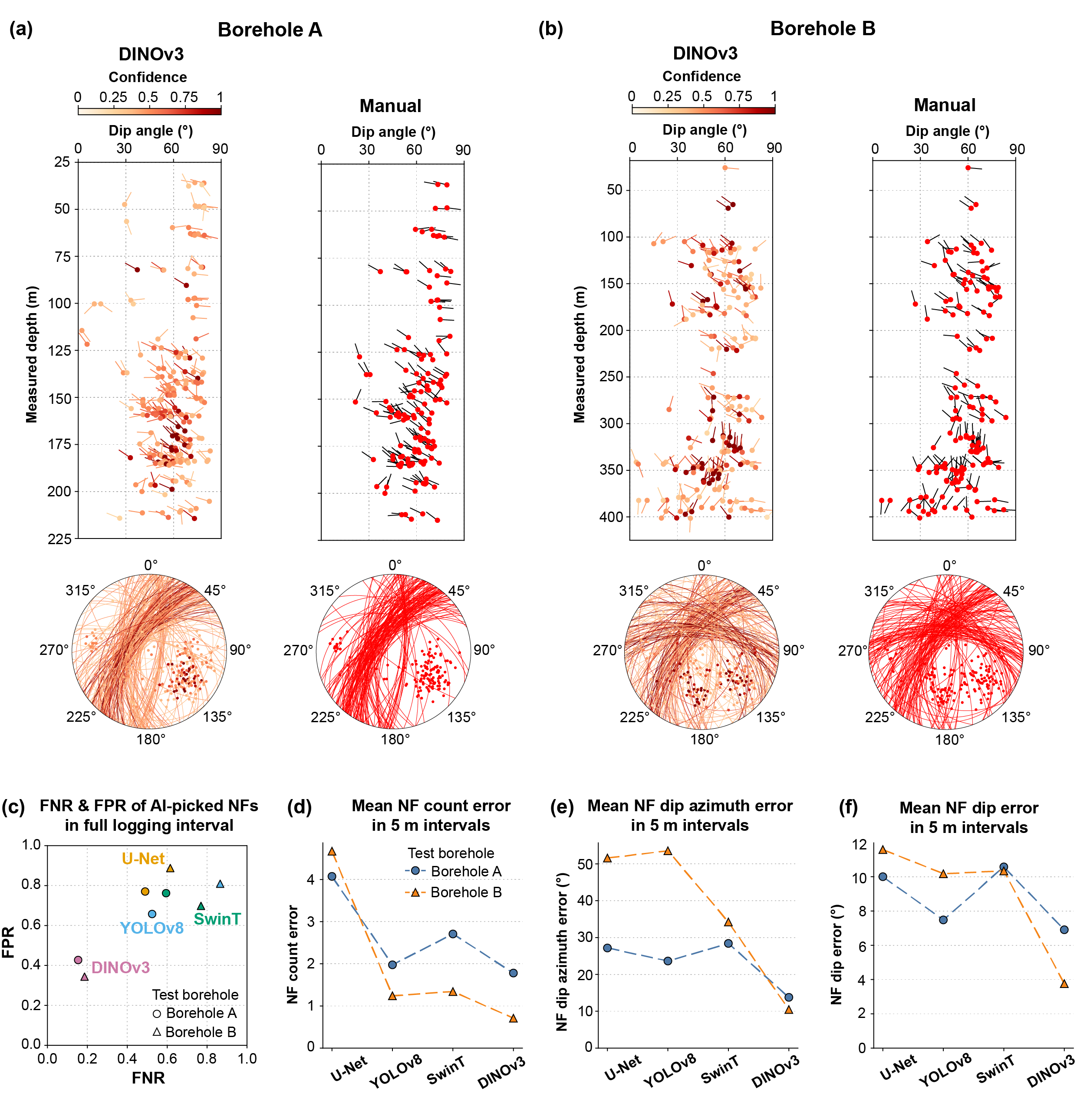}
\caption{NF characterization in Boreholes A and B. (a,b) DINOv3 and manual interpretations shown as tadpole and lower-hemisphere stereographic plots. Automated-pick shading denotes confidence. (c) Full-interval FPR and FNR of the picked NFs. (d--f) Mean absolute 5 m-window errors in fracture count, mean dip azimuth, and mean dip. }
\label{fig:nf_metrics}
\end{figure}

Across the full logs, confidence scores helps interpreters assess whether depth and orientation patterns are supported by higher-scoring picks or involve many lower-scoring candidates. With this visual guide, DINOv3's results show principal NF orientation patterns similar to those in the manual interpretations (Figures \ref{fig:nf_metrics}a and \ref{fig:nf_metrics}b). In Borehole A, the central dip azimuth and mean dip are about $310\degree$ and $60\degree$ for DINOv3, versus $309\degree$ and $61\degree$ for manual interpretation. In Borehole B, the corresponding values are $338\degree$ and $55\degree$ versus $338\degree$ and $57\degree$. This agreement in overall orientation does not imply that every individual pick matches the manual reference.

At the individual-pick level, DINOv3 misses fewer manually identified NFs and produces fewer unmatched automated picks than the models trained from scratch (Figure \ref{fig:nf_metrics}c). In Borehole A, its FNR is 0.15, compared with 0.49 for U-Net, the lowest FNR among the from-scratch models. This represents a relative reduction of about 69\%. DINOv3's FPR is 0.43, compared with 0.66 for YOLOv8, the lowest FPR among the from-scratch models. The same pattern holds in Borehole B, where DINOv3 has an FNR of 0.19 and an FPR of 0.34, both lower than those of any model trained from scratch.

DINOv3 also gives the lowest mean errors in NF count, dip azimuth, and dip across 5 m intervals in both boreholes (Figures \ref{fig:nf_metrics}d--f). The improvement is especially clear in Borehole B, where its count error is 0.71, 42.7\% below YOLOv8's 1.24, and its dip error is $3.76\degree$, compared with $10.17\degree$ for YOLOv8. For dip azimuth, the errors are $13.78\degree$ in Borehole A and $10.47\degree$ in Borehole B, below the lowest from-scratch values of $23.64\degree$ for YOLOv8 and $34.24\degree$ for Swin Transformer, respectively.

\subsection{Drilling-induced tensile fracture characterization}
\begin{figure}
\centering\figimage{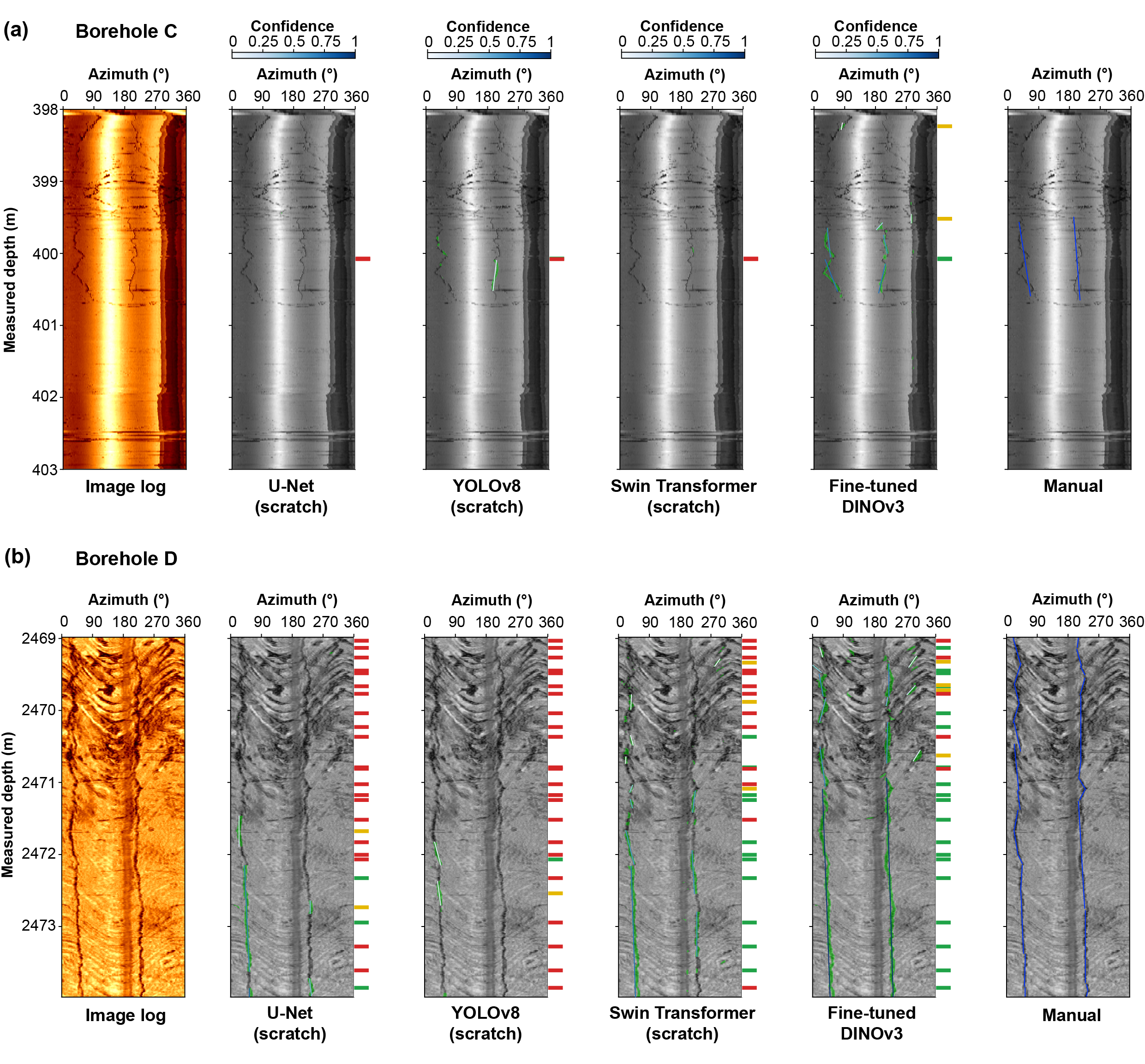}
\caption{DITF picking in selected 5 m intervals of (a) Borehole C and (b) Borehole D. Predicted masks and fitted centerlines show the recovered structures. Trace intensity indicates confidence. Red, yellow, and green side markers denote missed manual, unmatched automated, and matched picks, respectively.}
\label{fig:ditf_examples}
\end{figure}

We evaluate DITF picking along two other continuous logs, designated Boreholes C and D. Confidence scores also guide the review of individual DITF candidates by indicating how well their depth persistence and pairing match the expected geometry. Figure \ref{fig:ditf_examples} also uses darker fitted traces for higher confidence scores and lighter traces for lower scores, helping interpreters prioritize the more plausible candidates and identify lower-scoring picks for closer inspection.

The selected interval from Borehole C (Figure \ref{fig:ditf_examples}a) shows the effects of tool eccentering and logging noise, including a prominent dark band along the right edge of the image. Despite this interference, DINOv3 recovers more of the manually interpreted DITF pair than the models trained from scratch, although some unmatched picks remain. In the selected interval from Borehole D (Figure \ref{fig:ditf_examples}b), geological bedding creates curved and oblique features that complicate DITF picking. DINOv3 follows more of the two manually interpreted axial traces than the comparison models, but also detects additional segments.

\begin{figure}
\centering\figimage{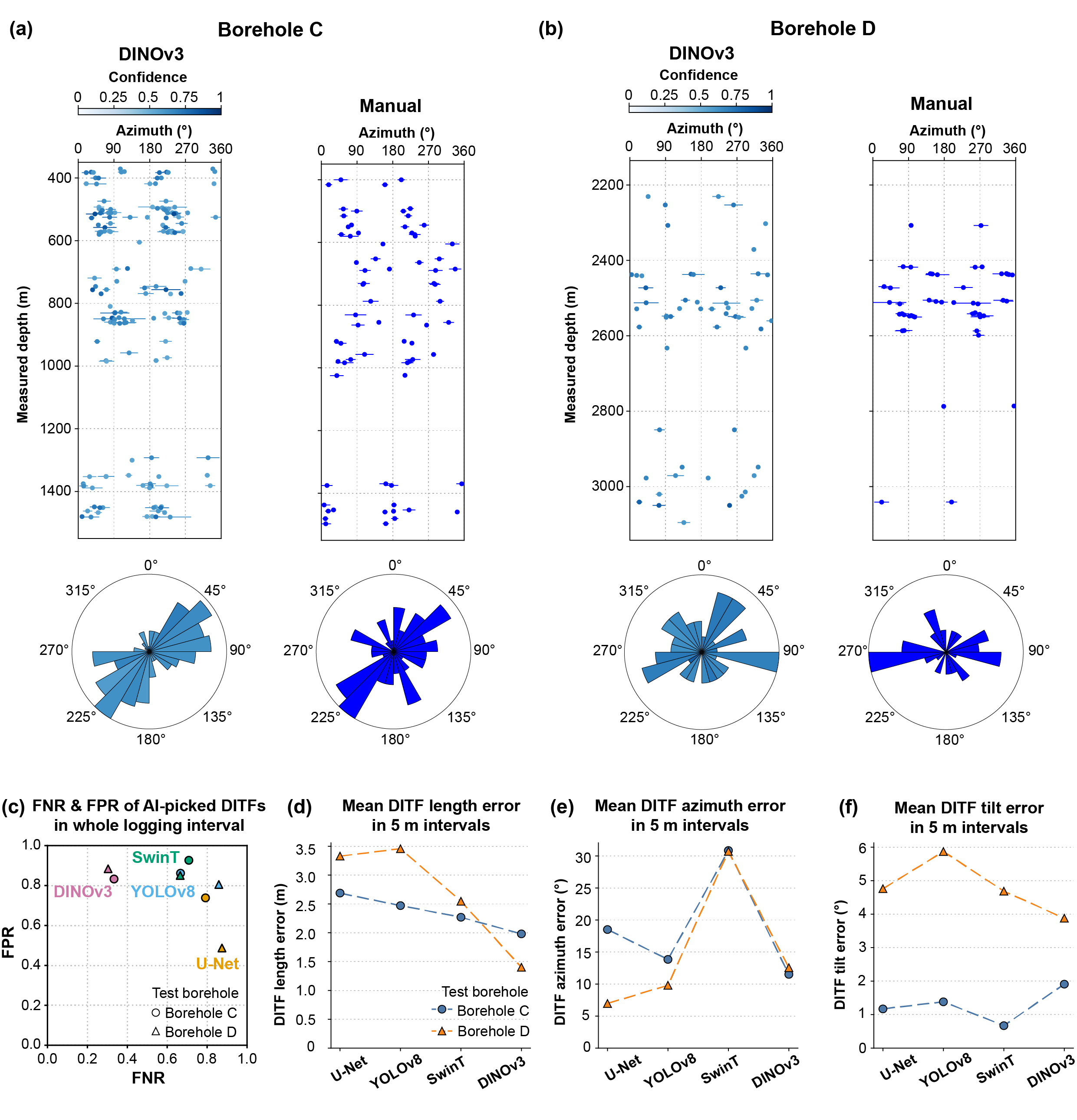}
\caption{DITF characterization in Boreholes C and D. (a,b) Depth--azimuth and rose diagram comparisons of DINOv3 and manual interpretations. Only DITFs with confidence scores higher than 0.5 are displayed. Bars represent DITF trace spans and dots represent central azimuths. (c) Full-interval FNR and FPR of the picked DITFs, and (d--f) mean absolute errors in DITF length,  mean azimuth,  and mean tilt within 5 m windows.  Note that panels (c--f) evaluate the full picking products rather than only the displayed high-confidence subset.}
\label{fig:ditf_metrics}
\end{figure}

For complete DITF logs, confidence scores also provide a way to limit the influence of potential false positives on the interpreted depth distributions and prevailing azimuths. We therefore display only DINOv3 picks with confidence greater than 0.5 in the depth--azimuth plots and rose diagrams (Figures \ref{fig:ditf_metrics}a and \ref{fig:ditf_metrics}b). Figures \ref{fig:ditf_unfiltered} shows the same comparisons without confidence score filtering. The threshold applies only to Figure \ref{fig:ditf_metrics}. The FNR, FPR, and parameter errors reported below are calculated from the full set of picks.

In both Boreholes C and D, the displayed DINOv3 picks closely reproduce the overall depth distribution and azimuth trends with depth of the manual picks. In Borehole C, the rose diagrams give a dominant DITF azimuth of $\mathrm{N}53\degree\mathrm{E}$ for DINOv3 and $\mathrm{N}45\degree\mathrm{E}$ for manual interpretation, with azimuthal standard deviations of $39\degree$ and $51\degree$, respectively. In Borehole D, the corresponding dominant azimuths are $\mathrm{N}90\degree\mathrm{E}$ and $\mathrm{N}88\degree\mathrm{E}$, with standard deviations of $63\degree$ and $49\degree$. The close agreement in DITF azimuths suggests that stress orientations inferred from DINOv3 picks would be consistent with those inferred manually.

At the individual-pick level, DINOv3 misses fewer manually identified DITFs than the models trained from scratch (Figure \ref{fig:ditf_metrics}c). Its FNR is 0.33 in Borehole C, compared with 0.67 for YOLOv8, the lowest from-scratch FNR. In Borehole D, DINOv3's FNR is 0.30 versus 0.67 for Swin Transformer. However, this improvement in detection completeness is accompanied by many unmatched candidates. DINOv3's FPR remains 0.83 in Borehole C and 0.88 in Borehole D, which is not the lowest in either borehole.

Beyond the number of detected DITFs, we assess how accurately the picks describe their total length and orientation within 5 m intervals. DINOv3 gives the lowest mean error in total DITF length in both boreholes (Figure \ref{fig:ditf_metrics}d). Its errors of 1.98 m in Borehole C and 1.40 m in Borehole D are 12.8\% and 45.1\% below Swin Transformer, the best from-scratch model for length in both boreholes. Orientation estimates are less consistently improved (Figures \ref{fig:ditf_metrics}e and \ref{fig:ditf_metrics}f). DINOv3 has the lowest azimuth error in Borehole C and the lowest tilt error in Borehole D, but U-Net gives a smaller azimuth error in Borehole D, at $6.96\degree$ versus $12.55\degree$ for DINOv3. Its tilt error in Borehole C is also higher than those of the three from-scratch models. Thus, DINOv3's advantage in detecting DITFs and estimating their total length does not extend to every orientation measure.

\FloatBarrier
\section{Discussion}
Using VFM adaptation in practice requires understanding both the basis of its performance and the choices that affect it when labeled data are limited. Therefore, we first examine whether DINOv3 responds to fracture traces before fine-tuning. Then, we assess how sample size, decoder design, fine-tuning strategy, and image enhancement affect performance. These analyses help identify suitable adaptation settings and clarify how much their benefits depend on the available data. We also discuss the limits of the current evaluation and their implications for geological and stress interpretation.

\subsection{DINOv3 responses to fracture traces before fine-tuning}

\begin{figure}[pos=!b]
\centering
\includegraphics[width=0.9\linewidth]{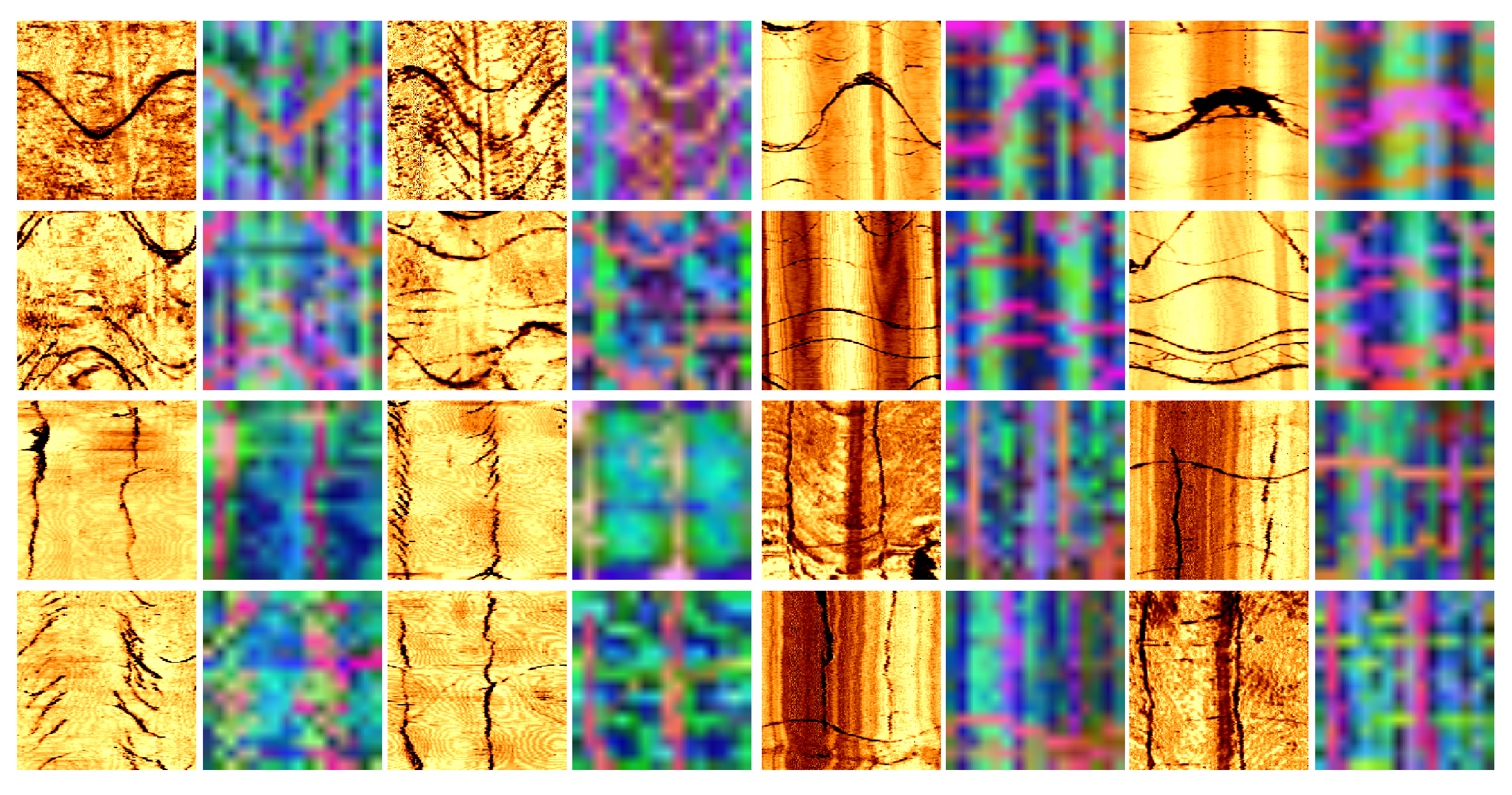}
\caption{DINOv3 responses to acoustic image logs before fine-tuning. PCA summarizes the responses at each image location in three values displayed as red, green, and blue. Within each image, similar colors generally indicate similar model responses, while contrasting colors indicate different responses. Examples are drawn from the validation and test sets. }
\label{fig:features}
\end{figure}

Fine-tuned DINOv3 generally outperformed the models trained from scratch. A possible explanation is its self-supervised pretraining on a large collection of satellite images. Although satellite images differ from acoustic image logs, pretraining may have enabled DINOv3 to learn visual patterns useful for identifying fractures. We therefore examine whether DINOv3 already responds to fracture traces in acoustic images before fine-tuning. Figure \ref{fig:features} displays these responses as color images. Principal component analysis (PCA) summarizes the model's responses at each image location in three values, which are displayed as red, green, and blue. Within each image, similar colors generally indicate similar model responses, whereas contrasting colors highlight differences. A fracture trace with a different color from the surrounding background therefore indicates that the model responds differently to these regions. No color is assigned a fixed meaning such as NF, DITF, or background, and the colors do not represent confidence scores.

In several examples, curved bands in the color images follow the sinusoidal NF traces, while near-vertical bands coincide with DITFs. These responses appear in images with different background textures and brightness patterns. Their clarity varies across examples, and the color bands are broader and less detailed than the original fracture traces. Background regions also produce changes in color. The pretrained model therefore responds to some fracture shapes, but these responses do not by themselves identify which structures should be labeled as NFs or DITFs.

A possible reason for these responses is that satellite images and acoustic image logs share visual patterns such as edges, elongated structures, and changes in texture. Learning these patterns during pretraining may give DINOv3 a useful starting point for locating fractures, even though satellite images do not teach it the geological meaning of an acoustic fracture trace. Fine-tuning can then link the existing responses to the NF and DITF labels and adjust them to the appearance of fractures in acoustic images. This provides a possible explanation for how the model can benefit from a small labeled dataset, since it can build on patterns learned before the fracture annotations are introduced.

\FloatBarrier
\subsection{Influence of training sample size}
\begin{figure}
\centering\figimage{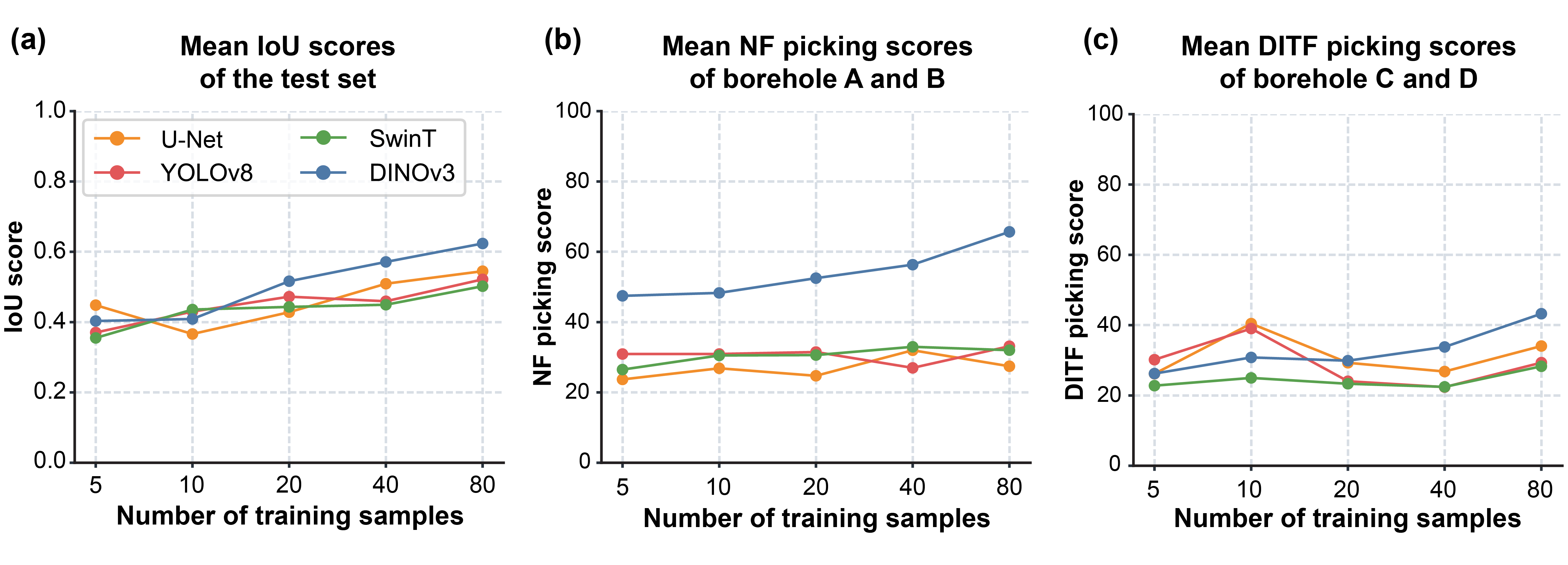}
\caption{Effect of training/fine-tuning sample amount on (a) mean IoU score for the test set, (b) mean NF picking score for Boreholes A and B, and (c) mean DITF picking score for Boreholes C and D. Picking scores combine detection and parameter accuracy as defined in Eq.~\ref{eq:composite}. }
\label{fig:sample_budget}
\end{figure}

To assess whether DINOv3's advantage persists with fewer than 80 labeled images, we repeat the comparison using smaller training sets. Reducing the number of training images affects segmentation and fracture picking differently (Figure \ref{fig:sample_budget}). For segmentation (Figure \ref{fig:sample_budget}a), DINOv3 gives the highest mean test IoU from 20 samples onward, but U-Net and Swin Transformer perform better with five and 10 samples, respectively. With 40 images, DINOv3 achieves an IoU of 0.571, exceeding the best result obtained by a model trained from scratch with 80 images, U-Net's 0.545. This comparison shows that DINOv3 can achieve better segmentation with half as many labeled images, although its advantage does not persist at the smallest sample sizes.

DINOv3's advantage is more consistent for NF picking, where it leads at every tested sample size (Figure \ref{fig:sample_budget}b). With only five labeled images, its score reaches 47.47, exceeding the scores of all three models trained from scratch with 80 images, which are no higher than 33.14. This differs from the segmentation ranking at the smallest sample size because IoU and fracture picking measure different aspects of the predictions. IoU is sensitive to trace thickness and boundary placement, whereas fracture depths and orientations depend on the position and shape of the extracted centerlines. A model can therefore produce useful NF measurements without achieving the highest IoU.

For DITF picking, DINOv3's advantage is less consistent when very few labeled images are available (Figure \ref{fig:sample_budget}c). YOLOv8 performs better with five images, and both U-Net and YOLOv8 perform better with 10. DINOv3 leads from 20 samples onward. DITF traces vary in shape and can be difficult to distinguish from noise, so adapting the model may require enough examples to cover this variation. The picking scores should also be read alongside the missed-pick fractions and parameter errors, because favorable parameter errors can partly offset missed detections in the combined score.

These results show that the advantage of VFM adaptation does not necessarily grow as the size of dataset becomes smaller. With very few training images, DINOv3's advantages in segmentation and DITF picking weaken or disappear, although its NF picking advantage persists. The irregular changes in some curves may reflect differences in the selected images and variation during training.  Nevertheless, within the tested range, larger training sets generally improve DINOv3's performance and make its advantage over models trained from scratch clearer. These results support VFM adaptation as an effective approach when labeled data are limited, while showing that additional fracture annotations can further improve performance and strengthen its advantage over training from scratch.

\subsection{Influence of decoder and fine-tuning strategy}
\label{sec:adaptation}
Adapting DINOv3 requires choosing both how to turn its image responses into fracture regions and how much of the pretrained model to update. These choices matter with limited data because training must preserve useful patterns while learning the appearance of fractures in acoustic images. We select three established decoder designs that differ in which encoder outputs they use and how they combine them.

The progressive upsampling (PUP) decoder uses the final encoder output and enlarges it step by step to produce a fracture map. The multi-level feature aggregation (MLA) decoder combines outputs from several encoder layers before producing the final prediction. Both decoders follow the designs of \citet{zheng2021rethinking}. The dense prediction transformer (DPT)-based decoder also uses several encoder layers, but merges their outputs progressively as the prediction is enlarged \citep{ranftl2021vision}. These designs represent different ways of recovering image detail from the encoder output. Comparing them tests whether combining information from several layers improves fracture characterization over using the final layer alone.

\begin{figure}
\centering\includegraphics[width=\linewidth]{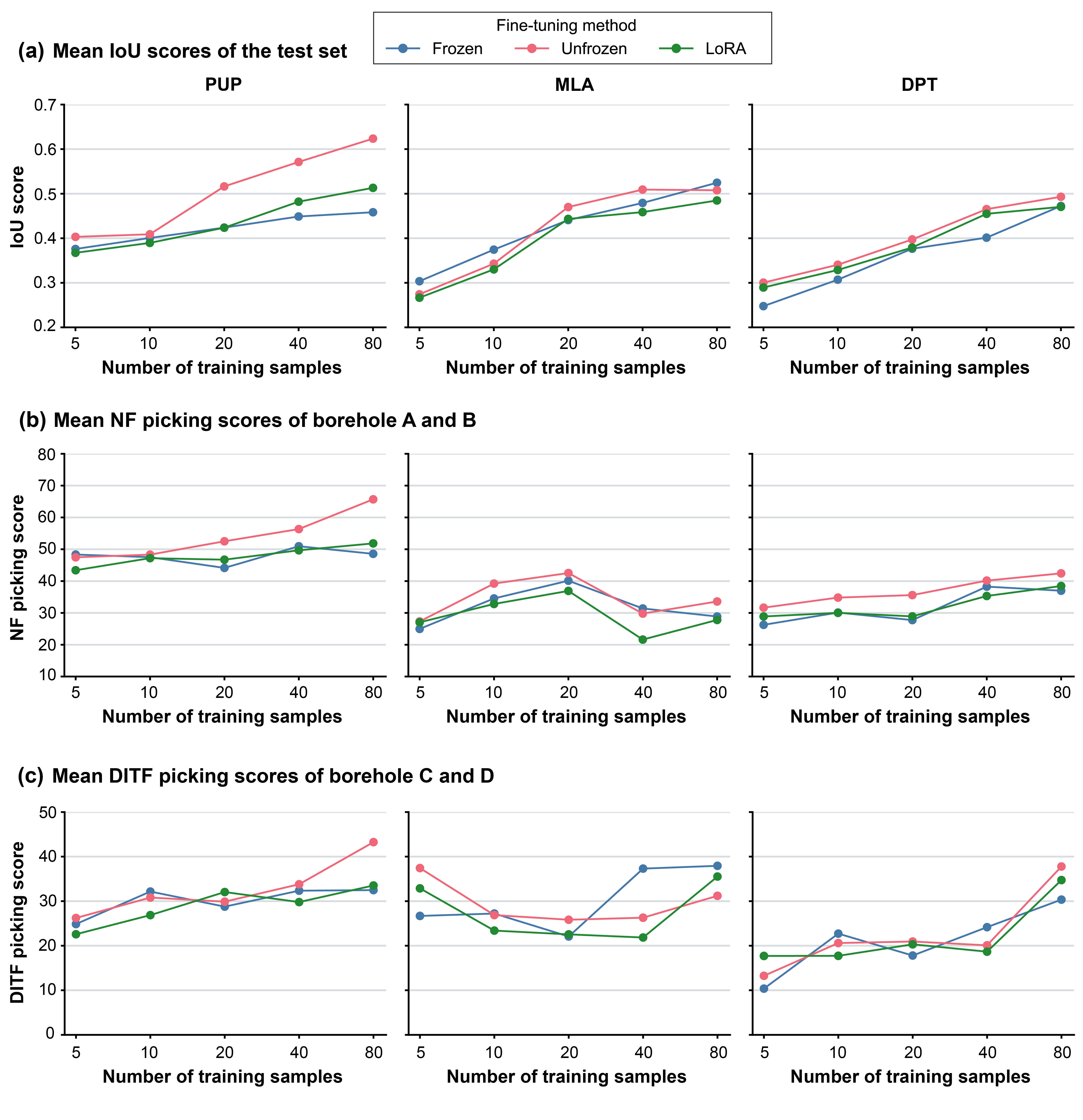}
\caption{Effects of decoder choice and fine-tuning strategy. Columns compare PUP, MLA, and DPT decoders. Curves compare Frozen, Unfrozen, and LoRA fine-tuning strategies. Rows show (a) mean IoU for the test set, (b) mean NF picking score for Boreholes A and B, and (c) mean DITF picking score for Boreholes C and D. }
\label{fig:adaptation}
\end{figure}

Each decoder is tested with three fine-tuning strategies (Figure \ref{fig:adaptation}). \emph{Frozen} keeps the pretrained encoder unchanged, whereas \emph{Unfrozen}, or full fine-tuning, updates the entire model. \emph{LoRA} keeps the encoder's original weights fixed and learns a small set of added adjustments \citep{hu2022lora}. The decoder is trained in every case, and all comparisons use the same pretrained encoder and numbers of labeled images.

PUP with full fine-tuning gives the strongest overall results with 80 training images. Its test IoU reaches 0.624, exceeding the best result from the other decoders (Figure \ref{fig:adaptation}a). It also gives the highest NF and DITF picking scores (Figures \ref{fig:adaptation}b and \ref{fig:adaptation}c). With the PUP decoder held constant, full fine-tuning outperforms both freezing and LoRA on all three measures at this sample size. Its benefit therefore extends from locating fracture pixels to obtaining useful fracture measurements.

Full fine-tuning is nevertheless not the best choice for every decoder and sample size. With only five training images, frozen PUP gives a slightly higher NF picking score than fully fine-tuned PUP. Frozen MLA also exceeds fully fine-tuned MLA in test IoU and DITF picking with 80 images. These exceptions suggest that updating more of the model can help or hinder performance depending on the decoder and the available examples. Similarly, a more complex decoder does not necessarily produce better fracture measurements. LoRA reduces the number of parameters updated during training, but this alone does not guarantee higher accuracy, and the present experiments do not compare training time or computing cost.

PUP gradually enlarges the encoder's output to produce a fracture map, while full fine-tuning allows the pretrained model to adjust to acoustic-image patterns. This combination may help retain fracture positions and shapes while learning to distinguish them from the background. The experiments show that limited labeled data do not automatically favor freezing the encoder or updating only a small part of it. For the tested dataset with 80 labeled images, the relatively simple PUP decoder combined with full fine-tuning gives the best overall fracture characterization.

\subsection{Effects of image enhancement on fracture characterization}
Fracture traces can be faint or obscured by background texture, suggesting that making them more visible might improve automated characterization. We therefore test whether image enhancement helps the models recover fractures and estimate their parameters more accurately. The processing combines automatic gain control (AGC), which adjusts local brightness and contrast, with ridge enhancement, which emphasizes narrow, elongated structures. The ridge-enhancement procedure includes the Sato line filter \citep{sato1998three}. For DINOv3, AGC-adjusted and ridge-filtered images replace the repeated copies of raw amplitude used in the original input. The comparison tests the combined processing, and each model uses the same processing during training and prediction.

\begin{figure}
\centering\figimage{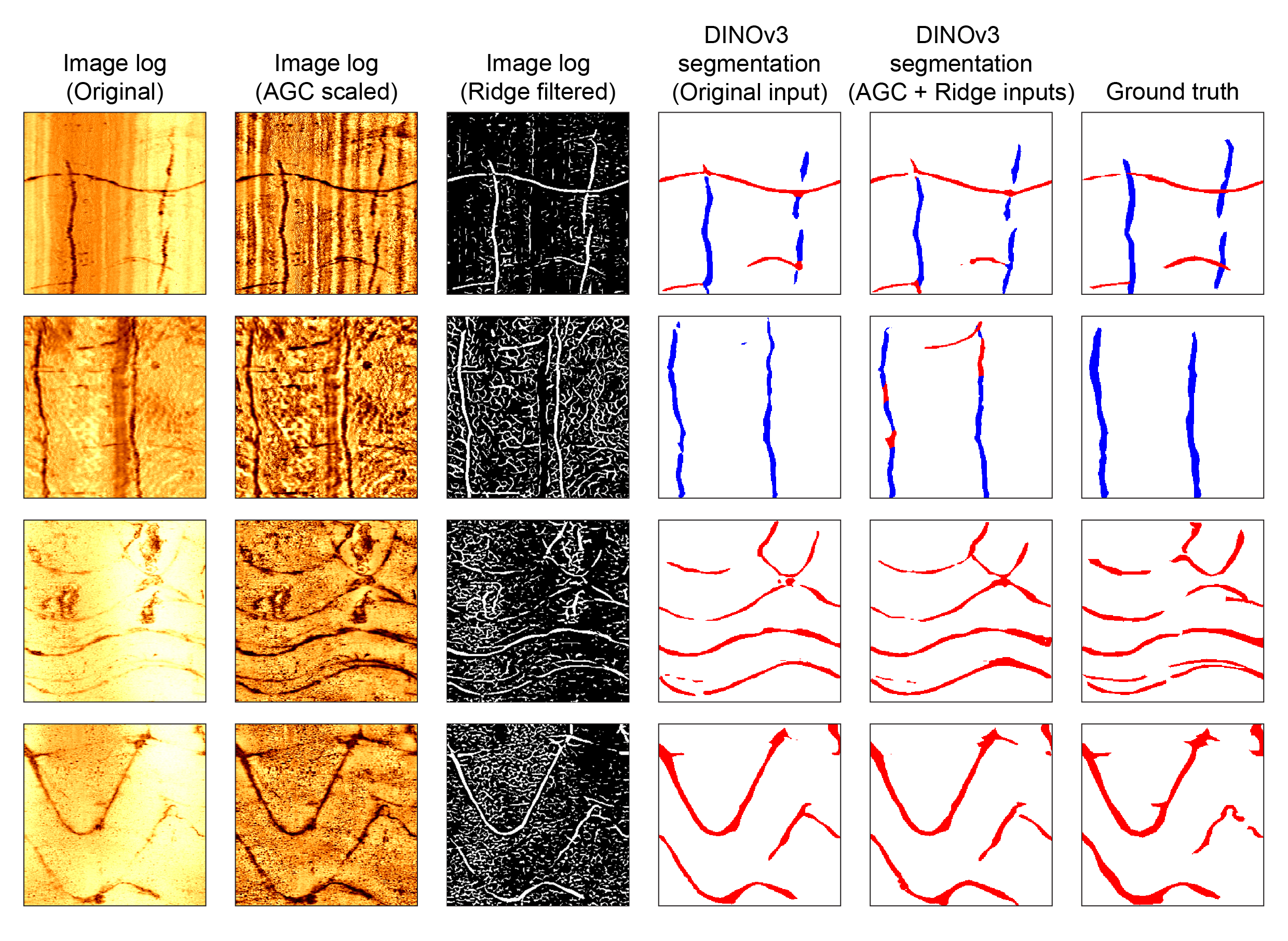}
\caption{Effect of image enhancement on DINOv3 fracture segmentation in selected test images. Columns show raw amplitude, the AGC-adjusted image, the ridge-filtered image, the prediction using raw amplitude, the prediction using the combined AGC and ridge-filtered images, and the manual labels. }
\label{fig:enhancement_examples}
\end{figure}

Figure \ref{fig:enhancement_examples} illustrates why a more visible fracture does not necessarily lead to a better prediction. Contrast adjustment can reveal faint traces, and ridge filtering can make narrow features more prominent. The same processing can also strengthen background features that resemble fractures or remove brightness differences that help distinguish fractures from their surroundings. These competing effects offer a possible explanation for the mixed prediction results, although the combined experiment does not separate the contributions of AGC and ridge filtering.

\begin{figure}
\centering\figimage{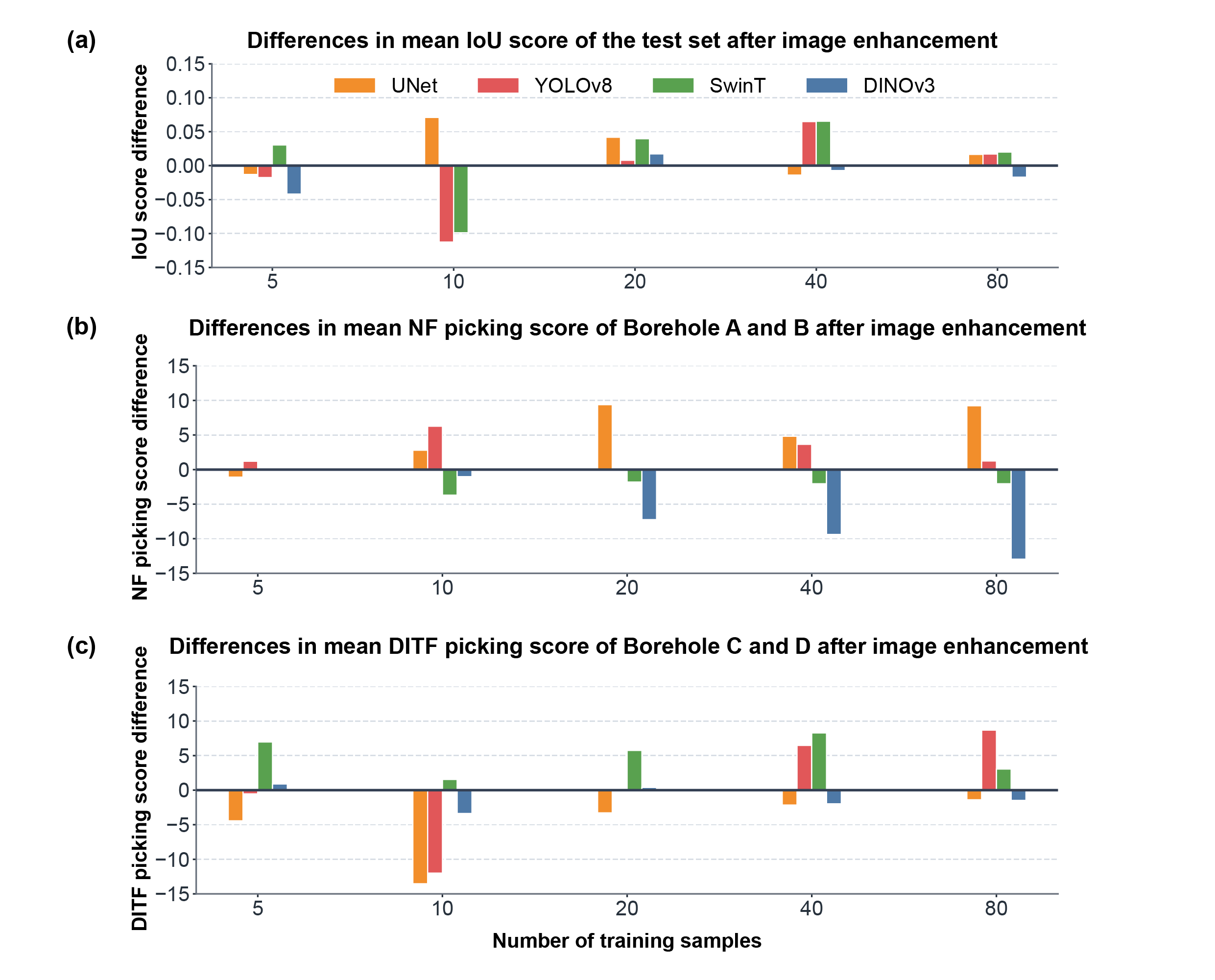}
\caption{Changes caused by image enhancement in (a) mean IoU score for the test set, (b) NF picking score for Boreholes A and B, and (c) DITF picking score for Boreholes C and D. Values are calculated as the enhanced-input result minus the raw-input result. Positive values favor enhancement and negative values favor raw amplitude. }
\label{fig:enhancement_metrics}
\end{figure}

The quantitative comparison confirms that enhancement is not consistently beneficial (Figure \ref{fig:enhancement_metrics}). With 80 training images, it slightly improves segmentation IoU for all three models trained from scratch, but DINOv3's IoU falls from 0.624 to 0.607 (Figure \ref{fig:enhancement_metrics}a). Even with enhancement, the models trained from scratch remain less accurate than DINOv3 using raw amplitude images. Enhancement does improve DINOv3's IoU with 20 training images, showing that its effect can change with sample size. Improving visual contrast therefore does not ensure better segmentation, even for the same model.

The fracture-picking results show the same dependence on the model and fracture type. With 80 training images, enhancement improves U-Net's NF picking score but reduces DINOv3's by 12.95 points (Figure \ref{fig:enhancement_metrics}b). For DITFs, YOLOv8 and Swin Transformer benefit, whereas DINOv3's score decreases (Figure \ref{fig:enhancement_metrics}c). Some enhanced U-Net experiments with very few training images produce no comparable DITF estimates for calculating certain errors. Their scores therefore include fewer measures and should not be interpreted as equivalent to scores calculated from every measure. For the tested data, the clearest practical choice is to fine-tune DINOv3 fully with the PUP decoder using raw amplitude images. Making fractures more visible through image enhancement does not guarantee more accurate picks and was not needed to obtain the best overall performance.

\subsection{Practical implications and limitations}
The main practical benefit is the ability to develop a fracture-characterization workflow from a small set of expert-labeled acoustic images. This offers a way to apply automated interpretation in projects without a large existing collection of fracture annotations. The improved NF picks and orientation estimates can support interpretation of rock mass structure along continuous logs. DITF picks can also help identify prevailing fracture azimuths for stress-orientation assessment, while confidence scores help prioritize candidates for review against the original images.

The usefulness of these outputs is nevertheless limited by what can be distinguished in the acoustic images. In particular, logging noise, tool eccentering, rough borehole walls, and geological bedding can obscure fractures or produce similar-looking patterns. As a result, some traces remain ambiguous even during expert review, and manual interpretations can also be uncertain. Agreement with manual picks therefore indicates consistency with an interpretation reference, but does not confirm the presence or absence of every physical fracture.

The parameter estimates also depend on simplified descriptions of fracture shape. A sinusoid summarizes an NF trace, and a straight line summarizes the overall trend of a DITF, but irregular or discontinuous traces may not follow these shapes closely. Image alignment, depth resolution, and borehole geometry further limit the precision of the fitted depths and orientations. Confidence scores inherit these geometric assumptions. For example, a true DITF can receive a low score if its opposite trace is not visible, while a background feature with a plausible shape may receive a higher score. Nevertheless, confidence scores remain useful for identifying more plausible fracture candidates and guiding their review against the original images.

The practical role of the workflow is therefore to assist geological and stress interpretation. It automates trace detection and fitting and supplies candidate fractures and parameter estimates for review. Expert judgment remains essential for ambiguous traces and for assessing the geological meaning of the resulting fracture patterns.

\FloatBarrier
\section{Conclusions}
This study investigated whether adapting a pretrained VFM improves fracture characterization in acoustic image logs compared with training task-specific models from scratch on the same limited labeled dataset. We adapted a DINOv3 model pretrained on satellite imagery and combined fracture segmentation with geometric processing to estimate NF and DITF depths and orientations.

With only 80 labeled images, fine-tuned DINOv3 achieved higher segmentation accuracy and more accurate NF detection, counts, and orientations than the models trained from scratch. For DITFs, detection and length estimates also improved, although unmatched picks remained common and orientation gains were less consistent. However, with very few labeled images, its segmentation and DITF picking advantages weakened or disappeared, whereas its NF picking advantage persisted. Further comparisons identified PUP with full fine-tuning using raw amplitude images as the best overall combination. In contrast, image enhancement did not consistently improve fracture characterization and was unnecessary for obtaining the best results.

Under the tested conditions, fine-tuning DINOv3 yielded better overall fracture characterization than training task-specific models from scratch on the same limited labeled dataset. These results support VFM adaptation as an effective strategy for acoustic image log interpretation with limited annotations. The resulting fracture measurements can support geological and stress analysis. However, expert review remains necessary, particularly for fractures with relatively low confidence scores. Finally, this study provides a useful reference for adapting VFMs to other geoscientific image analysis tasks with limited labeled data, particularly for selecting decoders and fine-tuning strategies.

\printcredits

\section*{Declaration of generative AI use}
During the preparation of this manuscript, the authors used ChatGPT
to improve language clarity and readability. All AI-assisted content was
reviewed and revised by the authors, who take full responsibility for the
manuscript.

\section*{Declaration of competing interest}
The authors declare that they have no known competing financial interests or personal relationships that could have appeared to influence the work reported in this paper.

\section*{Acknowledgments}
This work is financially supported by the National Science and Technology Major Project of China (Grant No. 2024ZD1000705).

\appendix
\section*{Appendix A. DITF picks without confidence score filtering}
\setcounter{figure}{0}
\renewcommand{\thefigure}{A\arabic{figure}}
\renewcommand{\theHfigure}{appendix.\thefigure}

\begin{figure}[pos=H]
\centering
\includegraphics[width=\linewidth]{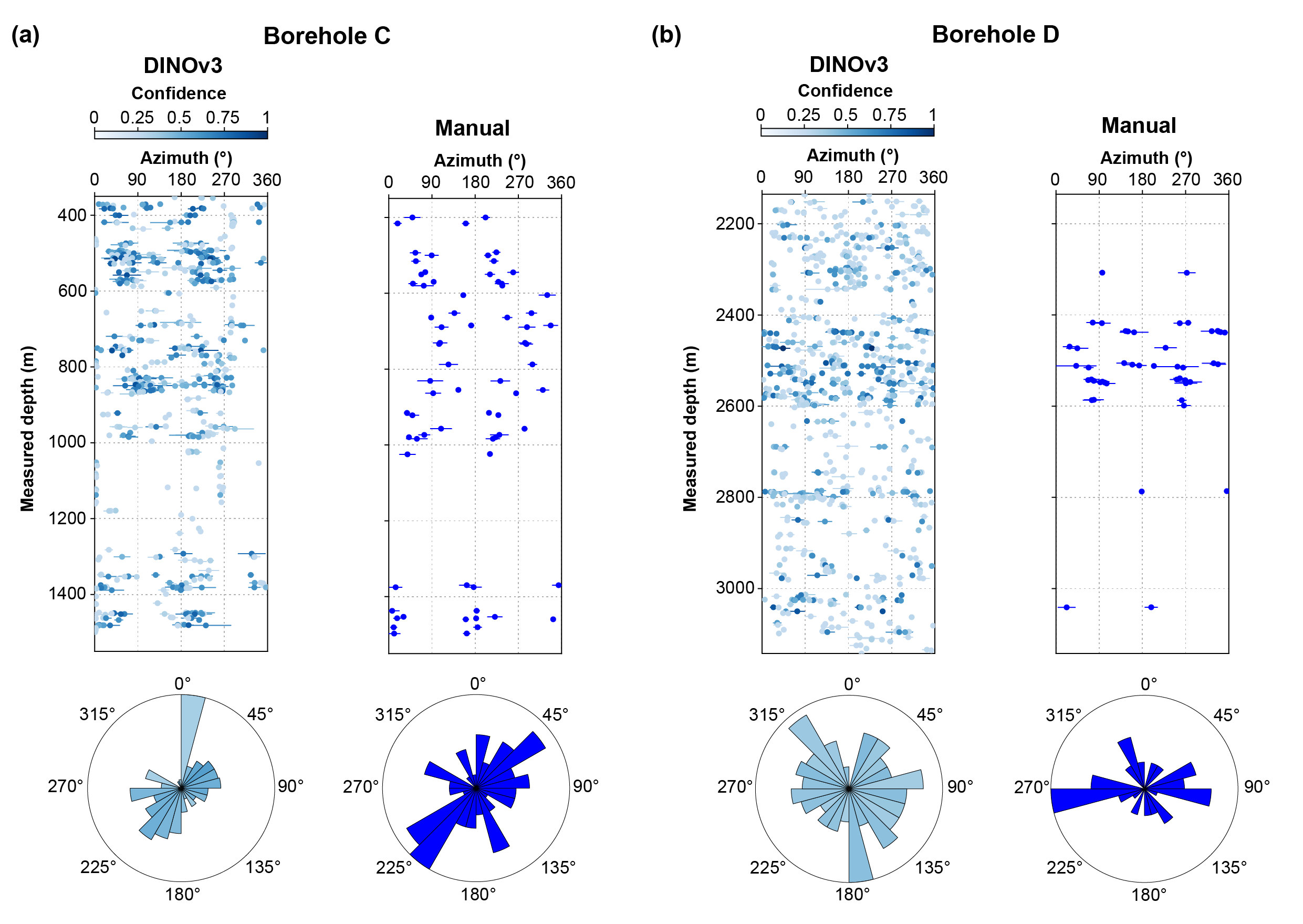}
\caption{Depth--azimuth plots and rose diagrams comparing DINOv3 and manual DITF picks in (a) Borehole C and (b) Borehole D. These panels correspond to Figures \ref{fig:ditf_metrics}a and \ref{fig:ditf_metrics}b but include all DINOv3 picks regardless of confidence score. }
\label{fig:ditf_unfiltered}
\end{figure}
\FloatBarrier

\section*{Data and code availability}
The source code and dataset used in this study will be made publicly available upon acceptance of the manuscript.

\bibliographystyle{cas-model2-names}
\bibliography{fracturepicker-refs}
\end{document}